\documentclass[english]{ccdconf}

\usepackage{graphics} % for pdf, bitmapped graphics files
\usepackage{epsfig} % for postscript graphics files
\usepackage{times} % assumes new font selection scheme installed
\usepackage{amsmath} % assumes amsmath package installed
\usepackage{amssymb}  % assumes amsmath package installed
\usepackage{cite}
\usepackage{cases}
\usepackage{subeqnarray}
\usepackage{array} % for better arrays (eg matrices) in maths
\usepackage{subfigure}
\usepackage{listings}
\usepackage{multirow}
\usepackage{cite}
\usepackage{cases}
\usepackage{subeqnarray}
\usepackage{etoolbox}
\usepackage{booktabs}
\usepackage{bm}

\DeclareMathOperator*{\minimize}{minimize}

\begin{document}

\title{Bi-objective Optimization for Robust RGB-D Visual Odometry}
\author{Tao Han\aref{amss}, Chao Xu\aref{amss}, Ryan Loxton\aref{hit}, Lei Xie\aref{amss}}

\affiliation[amss]{ The State Key Laboratory of Industrial Control Technology and the Institute of Cyber-Systems \& Control, Zhejiang University, Hangzhou 310027, China
        \email{ thancn@gmail.com, cxu@zju.edu.cn, leix@iipc.zju.edu.cn}}
\affiliation[hit]{ The Department of Mathematics \& Statistics, Curtin University, Perth 00301J, Australia
        \email{r.loxton@curtin.edu.au}}
\maketitle

\begin{abstract}
This paper considers a new bi-objective optimization formulation for robust RGB-D visual odometry. We investigate two methods for solving the proposed bi-objective optimization problem: the weighted sum method (in which the objective functions are combined into a single objective function) and the bounded objective method (in which one of the objective functions is optimized and the value of the other objective function is bounded via a constraint). Our experimental results for the open source TUM RGB-D dataset show that the new bi-objective optimization formulation is superior to several existing RGB-D odometry methods. In particular, the new formulation yields more accurate motion estimates and is more robust when textural or structural features in the image sequence are lacking.
\end{abstract}


%------------------------------------------------

\footnotetext{This work is supported by National High Technology Research and Development Program of China (863 Program) (2012AA041701), National Natural Science Foundation of China (61104048, 61473253).}

%=========================================================================================
\section{INTRODUCTION}

Visual odometry is an important area of information fusion in which the central aim is to estimate the pose of a robot using data collected by visual sensors \cite{nister2004visual}. Because nearly all robotic tasks require knowledge of the pose of the robot, visual odometry plays a critical role in robot control, simultaneous localization and mapping (SLAM) and robot navigation, especially when external reference information about the environment (such as GPS data) is unavailable. Visual odometry can be viewed as a particular instance of the general pose tracking problem, which is the most fundamental perception problem in robotics \cite{thrun2005probabilistic}.

To date, a variety of different visual odometry methods based on different sensor information have been studied and widely implemented. One of the most well-known methods is the iterative closest point (ICP) algorithm \cite{besl1992method}, which estimates the robot's pose by minimizing the distance between corresponding points in two laser scanning snapshots. However, this method can easily become trapped in local optima if a good initial guess is not provided. In addition to the ICP algorithm and its variants, odometry methods using camera images have also been studied \cite{henry2010rgb} \cite{strasdat2010scale}. Such methods usually extract point features from the camera images and match them through a series of steps, including descriptor matching, RANSAC and bundle adjustment. Due to their expensive computational burden, these approaches are usually too slow for real-time application. One way of improving computational efficiency is to use sparse point features, but this approach does not fully exploit the available image data, ignoring much relevant information.

Recently, with RGB-D cameras becoming smaller and cheaper, the opportunity has arisen to develop RGB-D odometry methods that exploit both intensity and depth information. One such method was proposed by the Computer Vision Group at the Technical University of Munich (TUM). In this method, a single-objective optimization problem is formulated to penalize the intensity difference between corresponding pixels in consecutive images \cite{steinbruecker_11iccv} \cite{kerl_13icra}. This method can be implemented in real-time even on a single-core CPU. However, the image depth information is only used to determine the relationship between corresponding pixels in consecutive images for intensity residual comparison; depth residuals are not considered. Thus, a new bi-objective optimization problem was subsequently proposed in \cite{tykkala2011direct} to minimize both depth and intensity residuals, with the aim of improving estimation performance.

In this paper, we consider the same bi-objective optimization formulation as in \cite{tykkala2011direct}. Our aims are twofold: (i) to propose new computational approaches for solving this bi-objective optimization formulation; and (ii) to explore and quantify the advantages of the bi-objective optimization formulation for improving estimation robustness. The first computational approach we investigate, the so-called weighted sum method, involves integrating the two objective functions into a single objective using a weighting factor. We derive a new formula for adaptive calculation of this weighting factor, which is crucial to estimation accuracy. Our formula is based on a novel \emph{image complexity} metric and differs from the corresponding formula in \cite{tykkala2011direct}, which uses the ratio of median intensity and median depth values to calculate the weighting factor. The second computational approach we investigate, the so-called bounded objective method, involves optimizing one of the objective functions while the other objective function is bounded via a constraint. Again, our new \emph{image complexity} metric is used, this time to determine an appropriate objective bound. To evaluate performance, the open source TUM RGB-D dataset \cite{sturm2012benchmark} was used. The computational results demonstrate that our new methods generally give results of superior accuracy compared with the methods in \cite{steinbruecker_11iccv} \cite{kerl_13icra} \cite{tykkala2011direct}.

%==========================================================================================
\section{SINGLE-OBJECTIVE OPTIMIZATION FOR VISUAL ODOMETRY}\label{single-objective optimization}

% 介绍 visual odometry 中传统的单目标优化模型
The camera motion in 3-D space has six degrees of freedom and can be denoted as
\begin{equation}\nonumber
\bm{\xi}=\left[ \nu_{1} ~ \nu_{2} ~ \nu_{3} ~ \psi_{1} ~ \psi_{2} ~ \psi_{3} \right]^{\top},
\end{equation}
where $\nu_{1}$, $\nu_{2}$, $\nu_{3}$ are the translation components of the motion and $\psi_{1}$, $\psi_{2}$, $\psi_{3}$ are the rotation components of the motion. To estimate $\bm{\xi}$, we consider a world point $\rho_{i}$ and assume that its brightness is the same in two consecutive images. This is the so-called photo-consistency assumption \cite{kerl_13icra}, which can be expressed mathematically by
\begin{equation}\nonumber
	I_{1}(\bm{x}_{i})=I_{2}(\bm{y}_{i}(\bm{\xi}^{*})),
\end{equation}
where $\bm{x}_{i} \in \mathbb{R}^{2}$ represents the mapping coordinate of the world point $\rho_{i}$ in the first image and $\bm{y}_{i}(\bm{\xi}^{*}) \in \mathbb{R}^{2}$ represents the corresponding coordinate of $\rho_{i}$ in the second image when given the true value of the camera motion $\bm{\xi}^{*}$. Moreover, $I_{1}(\cdot)$ and $I_{2}(\cdot)$ are the brightness (or intensity) values of the specified coordinates in the first and second images, respectively.

Based on the photo-consistency assumption, we can define the intensity difference corresponding to the motion estimate $\bm{\xi}$ as
\begin{equation}\nonumber
	r_{I}^{(i)}(\bm{\xi})=I_{2}(\bm{y}_{i}(\bm{\xi}))-I_{1}(\bm{x}_{i}).
\end{equation}
According to the results in \cite{kerl_13icra}, the more accurate the camera motion estimate, the smaller the residual $r^{(i)}_{I}(\bm{\xi})$. Thus, estimation quality in visual odometry can be assessed by considering the following least-squares objective function, which is the sum of residual squares for $n$ world points:
\begin{equation}\nonumber
	F_{I}(\bm{\xi})=\sum_{i=1}^{n}\left\{ r_{I}^{(i)}(\bm{\xi}) \right\}^{2}.
\end{equation}
Then the problem of determining the camera motion can be formulated as a least-squares optimization problem, i.e.,
\begin{equation}\label{single-objective optimization form}
	\minimize_{\bm{\xi}}~F_{I}(\bm{\xi}).
\end{equation}
To improve robustness, weighted residuals can be used to reduce the effect of noise and outliers in the image data. This motivates the following weighted objective function in quadratic form:
\begin{equation}\label{quadratic form}
	F_{I}(\bm{\xi})=\left[ \bm{r}_{I}(\bm{\xi}) \right]^{\top}\bm{\Omega}_{I}\left[ \bm{r}_{I}(\bm{\xi}) \right],
\end{equation}
where $\bm{\Omega}_{I}$ is a diagonal weight matrix and
\begin{equation}\nonumber
	\bm{r}_{I}(\bm{\xi})= \left[ r_{I}^{(1)}(\bm{\xi})~r_{I}^{(2)}(\bm{\xi})~\cdots~r_{I}^{(n)}(\bm{\xi}) \right]^{\top}.
\end{equation}

%==========================================================================================
\section{BI-OBJECTIVE OPTIMIZATION FOR RGB-D ODOMETRY}\label{bi-objective optimization}

\begin{figure}[t]\label{fig:1}
\centering
\includegraphics[width=\hsize]{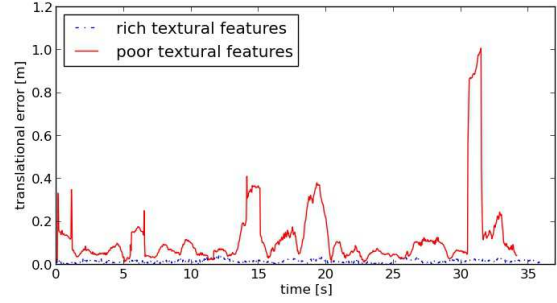}
\caption{Motion estimation accuracy of the single-objective Gauss-Newton method for the TUM RGB-D dataset.}
\end{figure}

Traditional cameras only provide image intensity information. RGB-D cameras, on the other hand, provide image intensity and image depth information, both of which can be used for visual odometry. For example, in the odometry methods introduced by the TUM Computer Vision Group \cite{steinbruecker_11iccv} \cite{kerl_13icra}, the relationship between corresponding pixels in consecutive images is expressed in terms of the depth information in the first image, and the intensity information of both images is used to define the motion estimation residuals as in Section \ref{single-objective optimization}. More precisely, the relationship between corresponding pixels in consecutive images is defined by a warping function as follows:
\begin{equation}\nonumber
	\bm{y}_{i}(\bm{\xi})=\bm{\tau}(\bm{\xi}, \bm{x}_{i}, D_{1}(\bm{x}_{i})),
\end{equation}
where $D_{1}(\bm{x}_{i})$ is the depth value of the pixel in the first image and $\bm{\tau}(\bm{\xi}, \bm{x}_{i}, D_{1}(\bm{x}_{i}))$ is the warping function for calculating the mapping coordinate $\bm{y}_{i}$ in the second image. For the specific form of the warping function $\bm{\tau}(\bm{\xi}, \bm{x}_{i}, D_{1}(\bm{x}_{i}))$, we refer the reader to \cite{kerl_13icra}.

Although single-objective optimization-based odometry methods are computationally fast and effective, they can produce poor results in some situations. For example, when textural features in the image sequence are poor, trajectory estimation accuracy will decrease dramatically. This is because the objective function $F_{I}(\bm{\xi})$ only depends on image intensity information, and thus it can become non-convex when image textural features are lacking. In this case, the ``optimal'' motion estimates obtained by applying an optimization iterative procedure may only be locally optimal. To investigate this hypothesis, we applied the single-objective optimization approach (implemented using the Gauss-Newton method) to image sequences in the TUM RGB-D dataset \cite{sturm2012benchmark}. Our results are shown in Fig. 1. From the results, we see that the translation error of the motion estimates increases significantly when textural features are lacking. This motivates the new bi-objective optimization formulation proposed in \cite{tykkala2011direct}, in which both image intensity and image depth residuals are minimized to improve robustness.

The extension of RGB-D odometry using bi-objective optimization is inspired by the ICP algorithm and its variants, which estimate the sensor motion by minimizing residual coordinate differences, instead of image intensity values. Since RGB-D cameras can provide both intensity and depth information simultaneously, we want to take full advantage of this feature by comparing depth differences, just as the ICP algorithm compares coordinate differences. Thus, we now consider two residuals instead of one:
\begin{equation}\label{new definition of residual}
	\begin{cases}
	&r^{(i)}_{I}(\bm{\xi})=I_{2}(\bm{\tau}(\bm{\xi}, \bm{x}_{i}, D_{1}(\bm{x}_{i})))-I_{1}(\bm{x}_{i}), \\
	&r^{(i)}_{D}(\bm{\xi})=D_{2}(\bm{\tau}(\bm{\xi}, \bm{x}_{i}, D_{1}(\bm{x}_{i}))) \\
	&~~~~~~~~~~~~~~~~~~~~~~  -[T(\bm{\xi}, \bm{x}_{i}, D_{1}(\bm{x}_{i}))]_{z},
	\end{cases}
\end{equation}
where $D_{1}(\cdot)$ and $D_{2}(\cdot)$ are the depth values of the specified coordinates in the first and second images, and $T(\bm{\xi}, \bm{x}_{i}, D_{1}(\bm{x}_{i}))$ projects the 3-D coordinate of world point $\rho_{i}$ from the first camera coordinate system to the second camera coordinate system based on the homogeneous transformation matrix for $\bm{\xi}$. Operator ``$[~]_{z}$'' selects the coordinate value along the $z$-direction. See the diagram in Fig. 2 for an explanation of the notation.

\begin{figure}[t]\label{fig:2}
\centering
\includegraphics[width=\hsize]{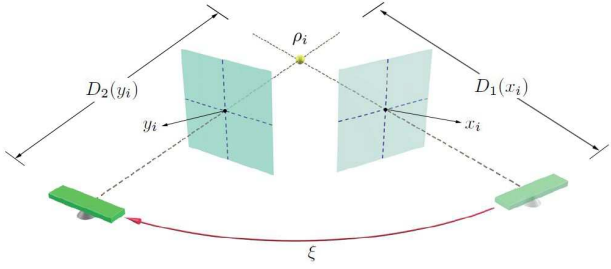}
\caption{Motion estimation via RGB-D odometry: $\rho_{i}$ is the world point under consideration, $x_{i}$ and $y_{i}$ are the pixels corresponding to $\rho_{i}$, and $D_{1}(x_{i})$ and $D_{2}(y_{i})$ are the depth values corresponding to $\rho_{i}$.}
\end{figure}

Based on $r^{(i)}_{D}(\bm{\xi})$ defined in \eqref{new definition of residual}, we consider the following objective function:
\begin{equation}\label{depth quadratic}
	F_{D}(\bm{\xi})=\left[ \bm{r}_{D}(\bm{\xi}) \right]^{\top}\bm{\Omega}_{D}\left[ \bm{r}_{D}(\bm{\xi}) \right],
\end{equation}
where $\bm{\Omega}_{D}$ is a diagonal weight matrix and
\begin{equation}\nonumber
	\bm{r}_{D}(\bm{\xi})= \left[ r_{D}^{(1)}(\bm{\xi})~r_{D}^{(2)}(\bm{\xi})~\cdots~r_{D}^{(n)}(\bm{\xi}) \right]^{\top}.
\end{equation}
Combining objectives \eqref{quadratic form} and \eqref{depth quadratic}, we consider the following bi-objective optimization problem:
\begin{equation}\label{multi-objective optimization}
	\minimize_{\bm{\xi}}~\bm{F}(\bm{\xi})=[F_{I}(\bm{\xi}),F_{D}(\bm{\xi})]^{\top}.
\end{equation}

\subsection{Weighted Sum Method}

The weighted sum method is the most common approach to solving multi-objective optimization problems. In this method, the individual objective functions are assigned different weights and then added together to form a single objective function. More specifically, for individual objective functions $\Psi_{1}, \Psi_{2}, \ldots, \Psi_{n}$ and decision vector $\bm{\alpha}$, the combined objective function is
\begin{equation}\label{weighted sum method}
	\Psi(\bm{\alpha})=\sum_{i=1}^{q}\omega_{i}\Psi_{i}(\bm{\alpha}),
\end{equation}
where $\omega_{i}$ are the weights. If all of the weights are positive, then the minimum of \eqref{weighted sum method} is Pareto optimal for the original multi-objective problem \cite{zadeh1963optimality}.

In essence, the objective weights provide additional degrees of freedom in the optimization problem. For our odometry problem \eqref{multi-objective optimization}, the new single-objective optimization problem is defined as
\begin{equation}\label{weighted sum method in our algorithm}
	\minimize_{\bm{\xi}}~F(\bm{\xi})=\omega_{I}F_{I}(\bm{\xi}) + \omega_{D}F_{D}(\bm{\xi}).
\end{equation}
Notice that by dividing $F(\bm{\xi})$ by $\omega_{I}$, we can obtain an equivalent optimization problem as follows:
\begin{equation}\label{new weighted sum form}
	\minimize_{\bm{\xi}}~\bar{F}(\bm{\xi})=F_{I}(\bm{\xi}) + \lambda F_{D}(\bm{\xi}),
\end{equation}
where $\lambda = \omega_{D} / \omega_{I}$. Thus, we only need to consider a single weighting factor $\lambda$.

Problem \eqref{new weighted sum form} can be solved using the Gauss-Newton method. To do this, we linearize the residuals $\bm{r}_{I}(\bm{\xi})$ and $\bm{r}_{D}(\bm{\xi})$ using the Taylor expansion proposed in \cite{kummerle2011g}:
\begin{equation}\nonumber
	\begin{cases}
	&\bm{r}_{I}(\bm{\xi} \oplus \Delta \bm{\xi}) \simeq \bm{r}_{I}(\bm{\xi}) + \bm{J}_{I}(\bm{\xi}) \Delta \bm{\xi}, \\
	&\bm{r}_{D}(\bm{\xi} \oplus \Delta \bm{\xi}) \simeq \bm{r}_{D}(\bm{\xi}) + \bm{J}_{D}(\bm{\xi}) \Delta \bm{\xi},
	\end{cases}
\end{equation}
where ``$\oplus$'' denotes the addition operator in Lie group SE(3) (for more details, see \cite{ma2003}); and $\bm{J}_{I}(\bm{\xi})$ and $\bm{J}_{D}(\bm{\xi})$ are the Jacobians defined by
\begin{equation}\nonumber
	\begin{split}
	\bm{J}_{I}(\bm{\xi})=\left.\frac{\partial \bm{r}_{I}(\bm{\xi} \oplus \Delta \bm{\xi})}{\partial\Delta\bm{\xi}}\right|_{\Delta\bm{\xi}=0}, \\
	\bm{J}_{D}(\bm{\xi})=\left.\frac{\partial \bm{r}_{D}(\bm{\xi} \oplus \Delta \bm{\xi})}{\partial\Delta\bm{\xi}}\right|_{\Delta\bm{\xi}=0}.
	\end{split}
\end{equation}
Then the objective function in \eqref{new weighted sum form} can be approximated by a quadratic function of $\Delta \bm{\xi}$:
\begin{equation}\label{quadratic form after linearisation}
	\begin{split}
	\bar{F}(\bm{\xi} \oplus \Delta \bm{\xi}) \simeq (a_{I} & + \lambda a_{D}) + 2(\bm{b}_{I}^{\top}+\lambda \bm{b}_{D}^{\top})\Delta\bm{\xi} \\
	                      & + \Delta\bm{\xi}^{\top}(\bm{H}_{I}+\lambda \bm{H}_{D})\Delta\bm{\xi},
	\end{split}
\end{equation}
where $a_{j}=\left[\bm{r}_{j}(\bm{\xi})\right]^{\top}\bm{\Omega}_{j}\bm{r}_{j}(\bm{\xi})$, $\bm{b}_{j}=\left[\bm{J}_{j}(\bm{\xi})\right]^{\top}\bm{\Omega}_{j}\bm{r}_{j}(\bm{\xi})$ and $\bm{H}_{j}=\left[\bm{J}_{j}(\bm{\xi})\right]^{\top}\bm{\Omega}_{j}\bm{J}_{j}(\bm{\xi})$ ($j=I,D$).

Suppose that at iteration $k$, we have the motion estimate $\bm{\xi}^{k}$. Then the increment $\Delta \bm{\xi}^{k}$ should be chosen to minimize $\bar{F}(\bm{\xi}^{k} \oplus \Delta \bm{\xi}^{k})$. According to the Gauss-Newton method, by differentiating \eqref{quadratic form after linearisation} for $\bm{\xi}=\bm{\xi}^{k}$, the optimal value of $\Delta \bm{\xi}^{k}$ satisfies the linear system
\begin{equation}\label{linear system}
	(\bm{H}_{I}^{k}+\lambda \bm{H}_{D}^{k})\Delta\bm{\xi}^{k} = -(\bm{b}_{I}^{k}+\lambda \bm{b}_{D}^{k}),
\end{equation}
where $\bm{b}_{j}^{k}$ denotes $\bm{b}_{j}$ with $\bm{\xi}=\bm{\xi}^{k}$ and $\bm{H}_{j}^{k}$ denotes $\bm{H}_{j}$ with $\bm{\xi}=\bm{\xi}^{k}$. To solve this linear system, methods such as Cholesky decomposition can be used. After solving \eqref{linear system}, the updated motion estimate is given by $\bm{\xi}^{k+1}=\bm{\xi}^{k} \oplus \Delta\bm{\xi}^{k}$. This iterative process continues until convergence is achieved.

The effectiveness of the weighted sum method depends crucially on the weighting factor $\lambda$, which must be selected a priori and reflects the preference of the decision maker. A good choice for $\lambda$ can result in more accurate trajectory estimates when compared to single-objective odometry methods, but a poor choice for $\lambda$ may lead to unacceptable results. Systematic approaches to selecting the weights in multi-objective optimization problems have been developed (see, for example, \cite{marler2004survey}), but few of them have been investigated in the context of visual odometry. Tykkala et al. \cite{tykkala2011direct} proposed a method that determines $\lambda$ based on the ratio of median intensity and median depth values:
\begin{equation}\nonumber
	\lambda=\big\vert \text{median}(I)/\text{median}(D) \big\vert^{2},
\end{equation}
where $I$ denotes the list of intensity values and $D$ denotes the list of depth values.

\begin{figure}[t]
\centering
\subfigure[Experiment 1 (rich structural features)]{
\label{fig:3_sub:1}
\includegraphics[width=\hsize]{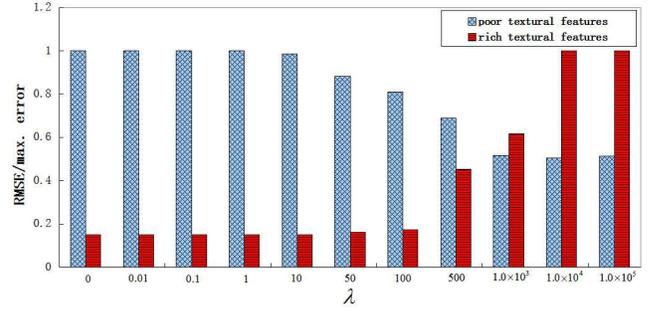}}
\subfigure[Experiment 2 (poor textural features)]{
\label{fig:3_sub:2}
\includegraphics[width=\hsize]{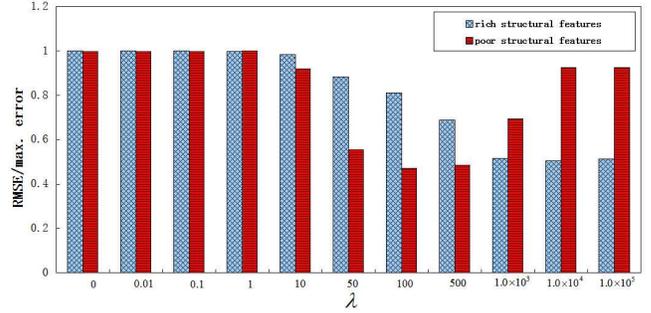}}
\caption{Ratio of root mean square error (RMSE) and maximum error for two computational experiments using the TUM RGB-D dataset.}
\label{fig:3}
\end{figure}

To explore the importance of the weight $\lambda$, we conducted two computational experiments with the TUM RGB-D dataset. For our first experiment, we formed two image sequences: one containing images with poor textural features and one containing images with rich textural features. The structural features in both image sequences were rich. We observed that for the first sequence with poor textural features, the error decreases as $\lambda$ is increased, but for the second sequence with rich textural features, the opposite occurs (see Fig. 3(a)). We believe that this is because the intensity objective function $F_{I}$ tends to be non-convex when images lack textural features. In this case, large values of $\lambda$ magnify the relative importance of the depth objective function $F_{D}$, thus potentially preventing the overall objective function in \eqref{new weighted sum form} from becoming non-convex.

For our second experiment, we again formed two image sequences: this time the first image sequence contained images with poor structural features and poor textural features, and the second image sequence contained images with rich structural features and poor textural features. As expected, the error decreases as $\lambda$ increases for the image sequence with rich structural features (see Fig. 3(b)). This is because $F_{D}$ is likely to be convex when images contain rich structural information, and a large $\lambda$ will increase $F_{D}$'s relative influence in the overall objective function.

Based on the experimental results in Fig. 3, we believe that the key to finding an optimal $\lambda$ is to design a metric to measure textural and structural information. To do this, we consider the concept of \emph{image complexity}, which is a measure of the inherent difficulty of finding a true target in a given image \cite{peters1990image}. Peters et al. \cite{peters1990image} has summarized many image complexity metrics for automatic target recognizers. Unfortunately, image complexity is a task-dependent notion and there is no universal metric applicable to all situations. After testing several of the metrics in \cite{peters1990image}, we designed our own metric for intensity complexity defined as follows:
\begin{equation}\label{image complexity metric}
	\begin{split}
	& \pi(I)= \dfrac{1}{(v-2)(h-2)} \sum_{i=2}^{v-1} \sum_{j=2}^{h-1} \left\{ \left\vert I(i+1, j) \right. \right. \\
	& ~~~ \left. -I(i-1, j) \right\vert + \left.\left\vert I(i, j+1)-I(i, j-1) \right\vert \right\},
	\end{split}
\end{equation}
where $v$ and $h$ are the number of pixel rows and pixel columns, respectively, and $I(\cdot, \cdot)$ denotes the intensity value at the specified pixel. For depth complexity, we use the analogue of \eqref{image complexity metric} for the depth values:
\begin{equation}\label{depth complexity metric}
	\begin{split}
	& \pi(D)= \dfrac{1}{(v-2)(h-2)} \sum_{i=2}^{v-1} \sum_{j=2}^{h-1} \left\{ \left\vert D(i+1, j) \right. \right. \\
	& ~~~ \left. -D(i-1, j) \right\vert + \left.\left\vert D(i, j+1)-D(i, j-1) \right\vert \right\},
	\end{split}
\end{equation}
where $D(\cdot, \cdot)$ denotes the depth value at the specified pixel. To standardize the intensity data $I$ and the depth data $D$, we define the following scaling factor as the ratio of the variance between them:
\begin{equation}\label{scale factor}
	\gamma=\dfrac{\sigma^{2}(I)}{\sigma^{2}(D)}.
\end{equation}
Combining \eqref{image complexity metric}-\eqref{scale factor}, we calculate the value of weight $\lambda$ as follows:
\begin{equation}\label{eq to calculate lambda}
	\lambda=\dfrac{\phi\gamma^{2}\pi(D)^{2}}{\pi(I)^{2}},
\end{equation}
where $\gamma$ is as defined in \eqref{scale factor} and $\phi$ is an adjustable constant. Notice that large values of $\pi(I)$ indicate rich textural features, and large values of $\pi(D)$ indicate rich structural features. Thus, we have deliberately chosen the value of $\lambda$ in \eqref{eq to calculate lambda} to be inversely proportional to $\pi(I)$, and proportional to $\pi(D)$. The idea is to use large values of $\lambda$ when the image sequence is rich in structure and/or poor in texture, and small values of $\lambda$ when the image sequence is poor in structure and/or rich in texture.

\subsection{Bounded Objective Method}

The bounded objective method is another method for solving multi-objective optimization problems \cite{marler2004survey}. In this method, we minimize one of the objective functions (considered as the most important, or primary, objective), while the other objective functions are bounded using additional constraints.

For our odometry problem, we select $F_{I}(\bm{\xi})$ as the primary objective function. The bi-objective optimization problem in \eqref{multi-objective optimization} then becomes
\begin{equation}\label{bounded objective form}
	\begin{split}
	\minimize_{\bm{\xi}} ~ &F_{I}(\bm{\xi}) \\
	\text{subject to} ~ &F_{D}(\bm{\xi}) \leq \epsilon_{D},
	\end{split}
\end{equation}
where $\epsilon_{D}$ is an upper bound for the least-squares sum of depth residuals. To solve the optimization problem in \eqref{bounded objective form}, we can again use the first-order Taylor expansions of $\bm{r}_{I}(\bm{\xi} \oplus \Delta \bm{\xi})$ and $\bm{r}_{D}(\bm{\xi} \oplus \Delta \bm{\xi})$. The optimal increment $\Delta \bm{\xi}$ at point $\bm{\xi}$ is then given by the solution of the following problem:
\begin{equation}\label{QCQP}
	\begin{split}
	\minimize_{\Delta\bm{\xi}} ~ &\Delta\bm{\xi}^{\top}\bm{H}_{I}\Delta\bm{\xi}+2\bm{b}_{I}^{\top}\Delta\bm{\xi}+a_{I} \\
	\text{subject to} ~ &\Delta\bm{\xi}^{\top}\bm{H}_{D}\Delta\bm{\xi}+2\bm{b}_{D}^{\top}\Delta\bm{\xi}+a_{D} \leq \epsilon_{D},
	\end{split}
\end{equation}
where $\bm{H}_{I}$, $\bm{H}_{D}$, $\bm{b}_{I}$, $\bm{b}_{D}$, $a_{I}$ and $a_{D}$ are as defined in \eqref{quadratic form after linearisation}.

\begin{figure*}
\centering
\subfigure[Poor structure $\&$ poor texture]{
\label{fig:4_sub:1}
\includegraphics[width=0.23\textwidth]{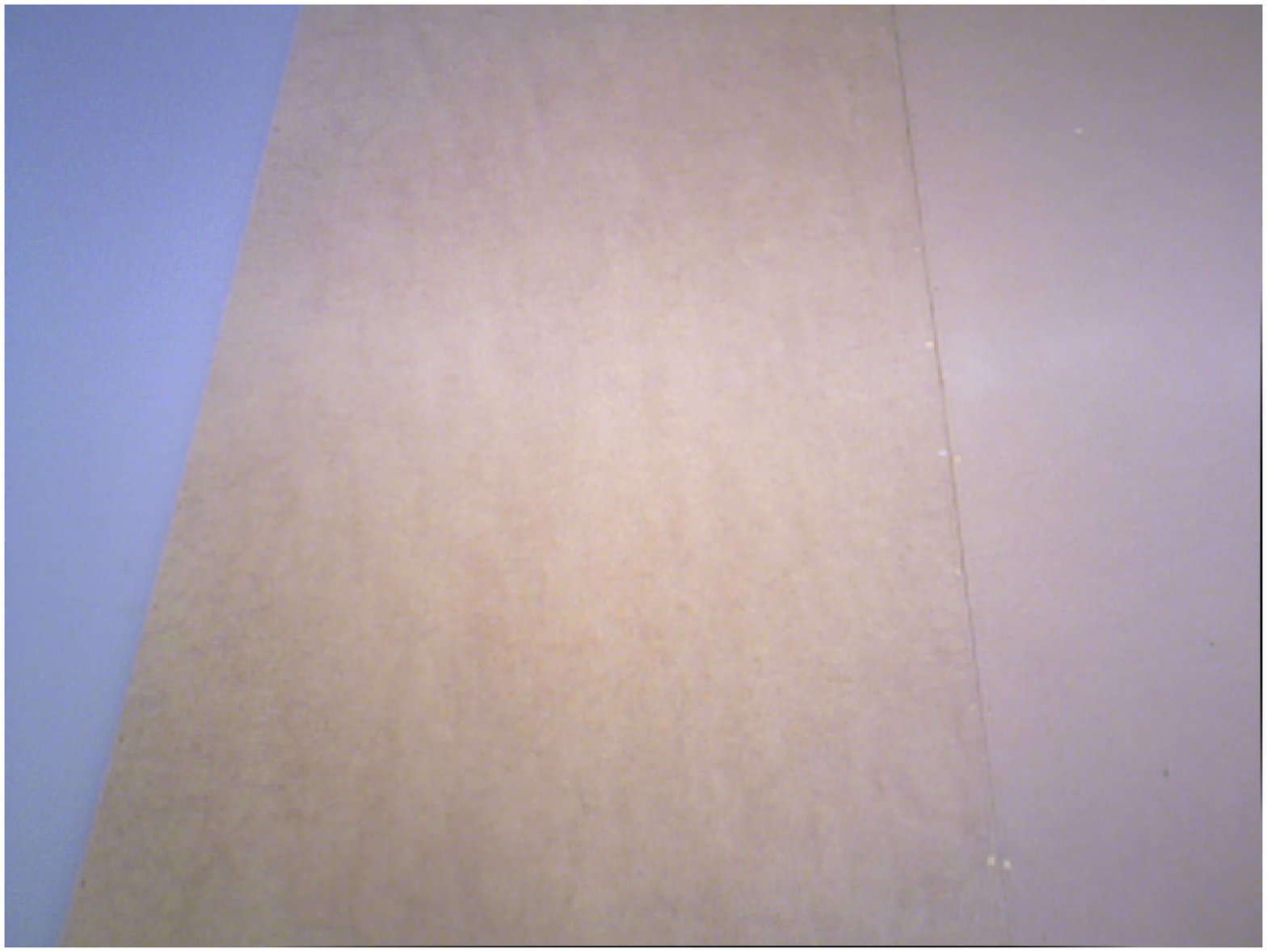}}
\subfigure[Poor structure $\&$ rich texture]{
\label{fig:4_sub:2}
\includegraphics[width=0.23\textwidth]{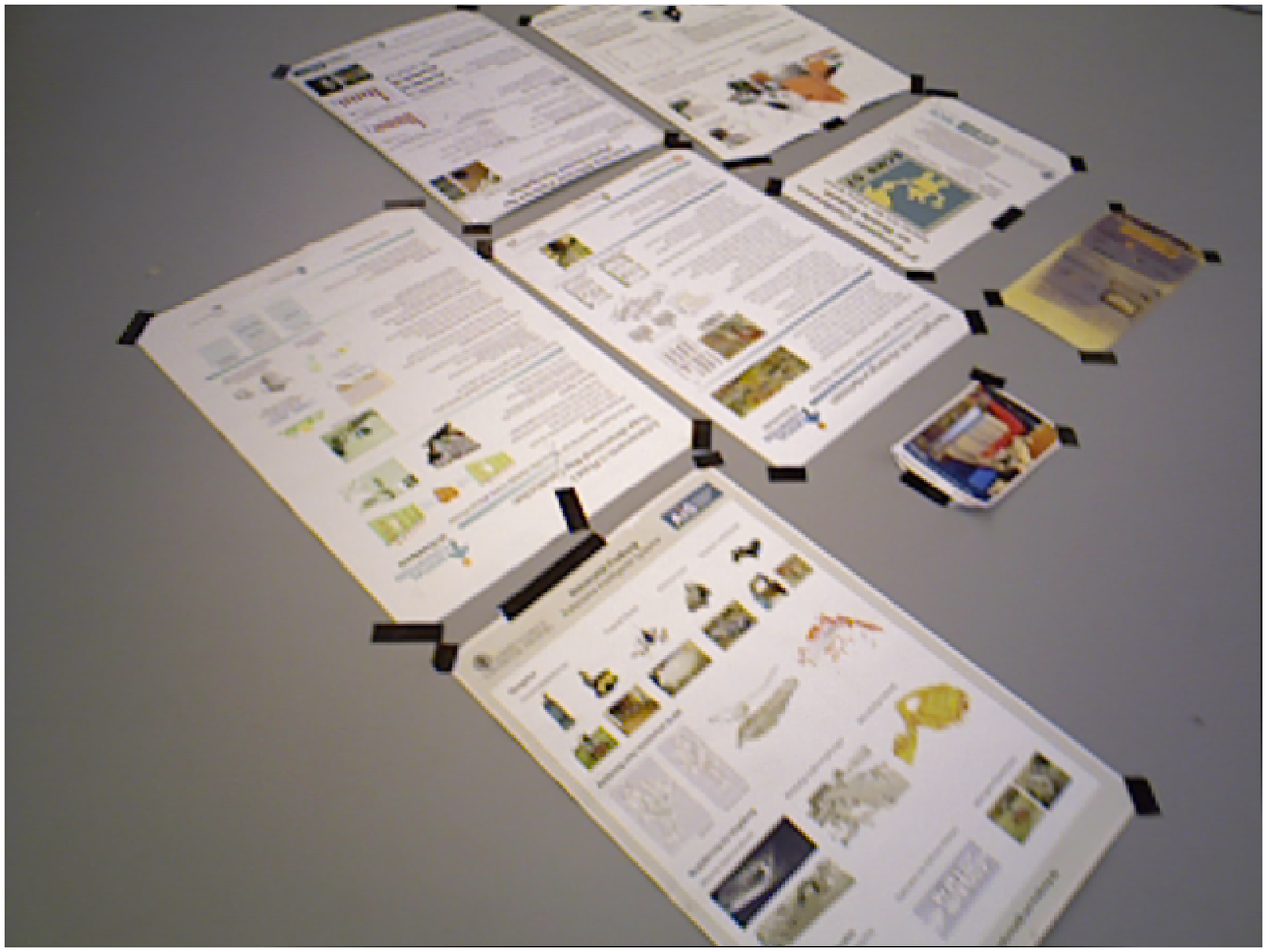}}
\subfigure[Rich structure $\&$ poor texture]{
\label{fig:4_sub:3}
\includegraphics[width=0.23\textwidth]{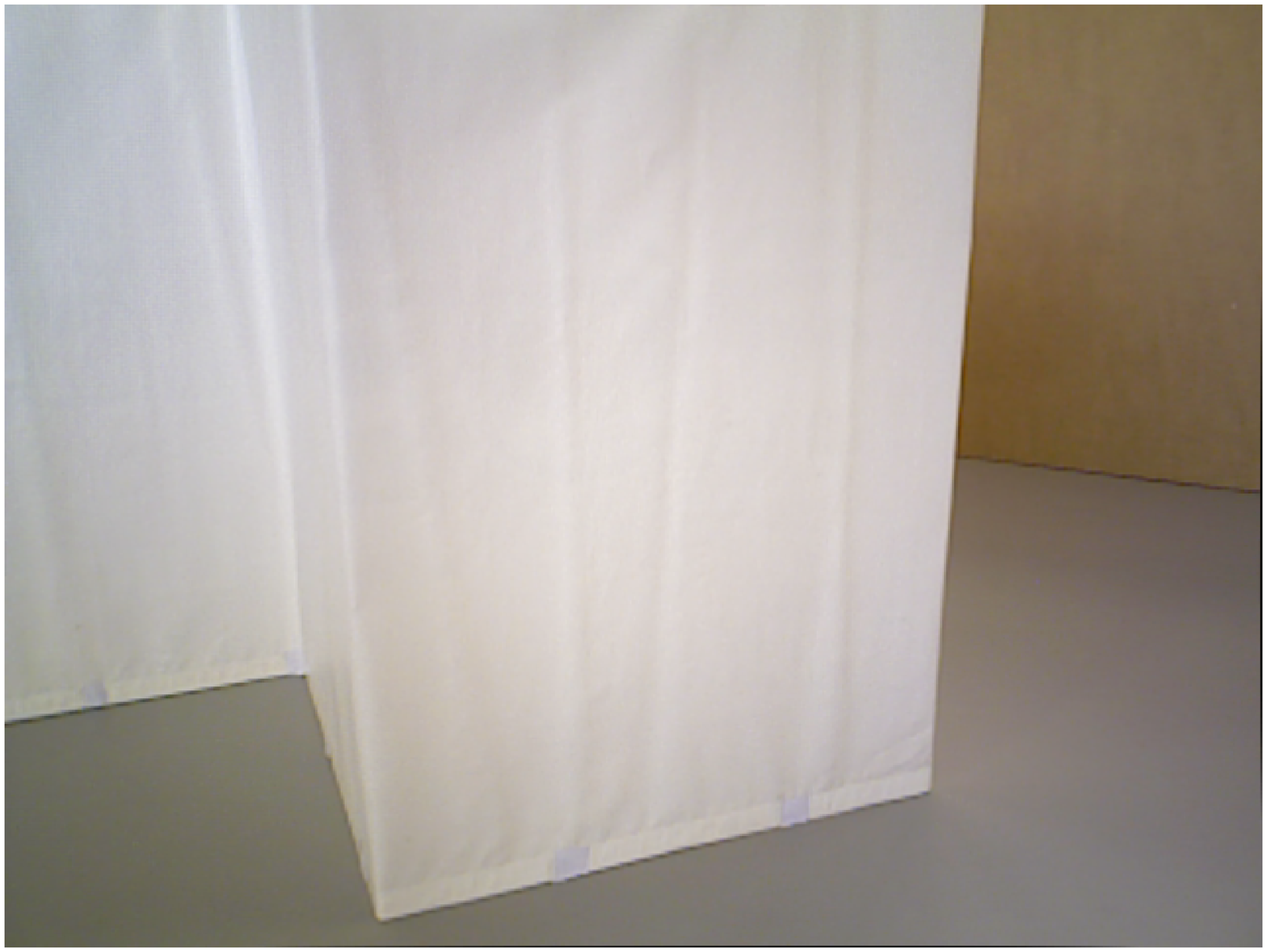}}
\subfigure[Rich structure $\&$ rich texture]{
\label{fig:4_sub:4}
\includegraphics[width=0.23\textwidth]{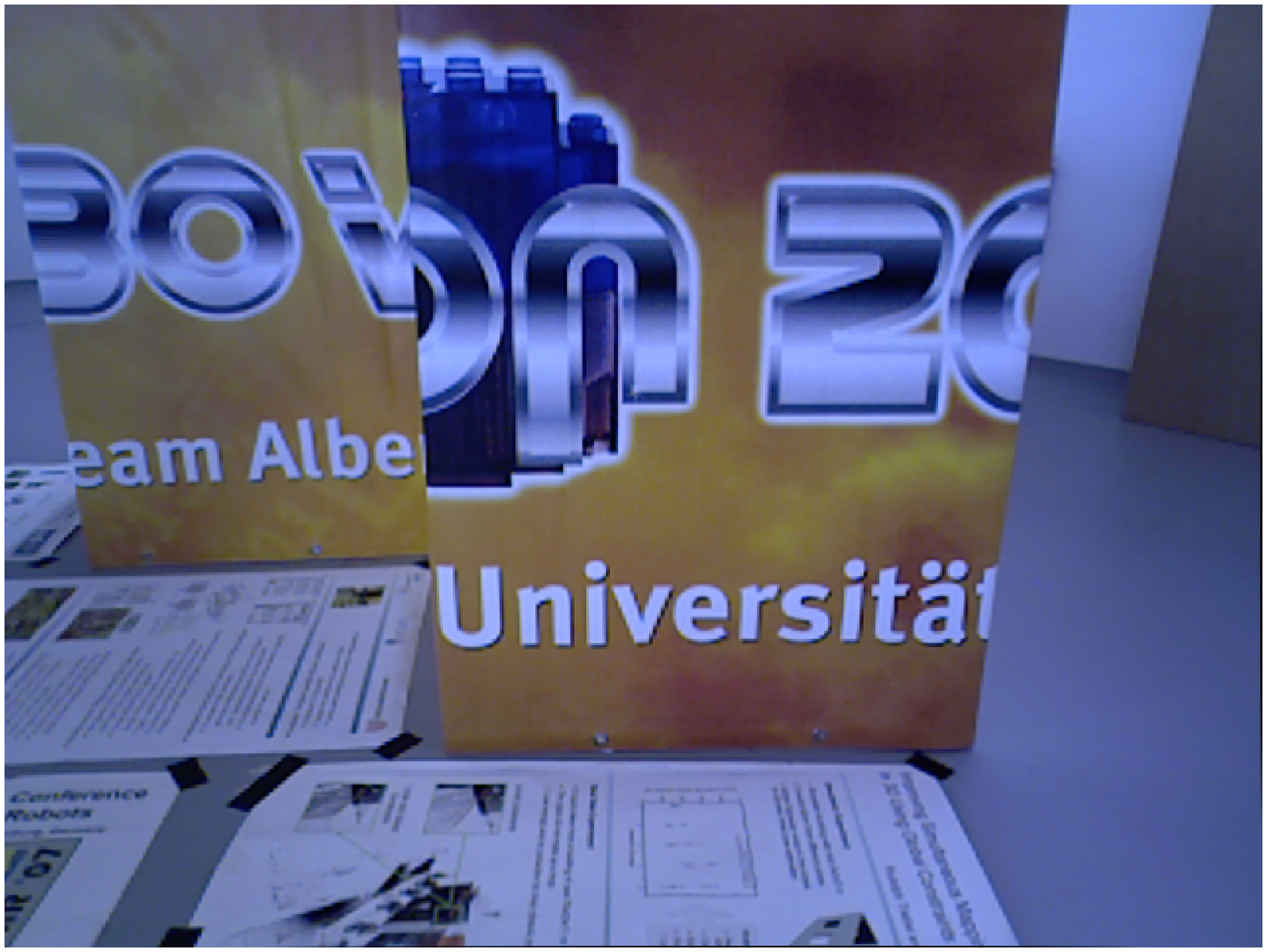}}
\caption{The four types of images in the ``Structure vs. Texture'' category in the TUM RGB-D dataset.}
\label{fig:4}
\end{figure*}

Problem \eqref{QCQP} is a \emph{quadratically constrained quadratic program} (QCQP). The general form for a QCQP is
\begin{equation}\nonumber
	\begin{split}
	\minimize_{\bm{\alpha} \in \mathbb{R}^{n}} ~ &\bm{\alpha}^{\top}\bm{H}_{0}\bm{\alpha}+2\bm{b}_{0}^{\top}\bm{\alpha}+a_{0} \\
	\text{subject to} ~ &\bm{\alpha}^{\top}\bm{H}_{i}\bm{\alpha}+2\bm{b}_{i}^{\top}\bm{\alpha}+a_{i} \leq 0,~i=1,\ldots,q.
	\end{split}
\end{equation}
QCQPs are of both theoretical and practical significance \cite{lu2011kkt}. Because the matrices $H_{I}$ and $H_{D}$ are positive semidefinite, problem \eqref{QCQP} is a convex QCQP. To solve this convex QCQP, we first transform it into a \emph{second-order cone programming} (SOCP) problem and then apply SOCP techniques \cite{lobo1998applications}. The general form for a SOCP problem is
\begin{equation}\nonumber
	\begin{split}
	\minimize_{\bm{\alpha} \in \mathbb{R}^{n}} ~ &\bm{c}^{\top}\bm{\alpha} \\
	\text{subject to} ~ & \left\Vert \bm{A}_{i}\bm{\alpha}+\bm{p}_{i} \right\Vert \leq \bm{q}_{i}^{\top}\bm{\alpha}+d_{i},~i=1,\ldots,q.
	\end{split}
\end{equation}
The norm appearing in the constraints is the standard Euclidean norm, i.e., $\left\Vert \bm{u} \right\Vert=(\bm{u}^{\top}\bm{u})^{1/2}$. We first rewrite \eqref{QCQP} as follows:
\begin{equation}\label{QCQP2}
	\begin{split}
	\minimize_{\Delta\bm{\xi}} ~ &\left\Vert \bm{\Omega}_{I}^{1/2}\bm{J}_{I}\Delta\bm{\xi} + \bm{\Omega}_{I}^{1/2}\bm{r}_{I} \right\Vert^{2} \\
	\text{subject to}~ &\left\Vert \bm{\Omega}_{D}^{1/2}\bm{J}_{D}\Delta\bm{\xi} + \bm{\Omega}_{D}^{1/2}\bm{r}_{D} \right\Vert^{2} \leqslant \epsilon_{D}.
	\end{split}
\end{equation}
By adding a new optimization variable $t \in \mathbb{R}$, we can transform \eqref{QCQP2} into the following SOCP form:
\begin{equation}\label{SOCP in our method}
	\begin{split}
	\minimize_{(\Delta\bm{\xi},t)} ~ &t \\
	\text{subject to}~ &\left\Vert \bm{\Omega}_{I}^{1/2}\bm{J}_{I}\Delta\bm{\xi} + \bm{\Omega}_{I}^{1/2}\bm{r}_{I} \right\Vert \leqslant t \\
								&\left\Vert \bm{\Omega}_{D}^{1/2}\bm{J}_{D}\Delta\bm{\xi} + \bm{\Omega}_{D}^{1/2}\bm{r}_{D} \right\Vert \leqslant \sqrt{\epsilon_{D}}.
	\end{split}
\end{equation}
Problem \eqref{SOCP in our method}, which is equivalent to \eqref{QCQP} and \eqref{QCQP2} (see \cite{lobo1998applications}), is clearly in the general SOCP form shown above.

To solve the SOCP problem in \eqref{SOCP in our method}, we can use ECOS, an SOCP solver developed by Domahidi et al. \cite{bib:Domahidi2013ecos}. ECOS implements an interior point method to solve SOCPs in the following standard form \cite{andersen2013cvxopt}:
\begin{equation}\nonumber
	\begin{split}
	\minimize_{\bm{\alpha} \in \mathbb{R}^{n}} ~ &\bm{c}^{\top}\bm{\alpha} \\
	\text{subject to} ~ &\bm{G}\bm{\alpha}+\bm{s}=\bm{h},~\bm{s} \in \bm{K},
	\end{split}
\end{equation}
where $\bm{\alpha}$ is a vector of optimization variables, $\bm{s}$ is a vector of slack variables and $\bm{K}$ is the cone
\begin{equation}\nonumber
	\bm{K}=\prod_{\mu = 1}^{N} \{ (u_{0},\bm{u}_{1}) \in \mathbb{R} \times \mathbb{R}^{m_{\mu} -1} ~ : ~ u_{0} \geq \left\Vert \bm{u}_{1} \right\Vert \}.
\end{equation}
To reformulate \eqref{SOCP in our method} into the standard form required by ECOS, we set
\begin{equation}\nonumber
\bm{\alpha}=\left[
	\begin{array}{lcr}
	~t \\
	\Delta\bm{\xi}
	\end{array}
	\right],
\end{equation}
and set
\begin{equation}\nonumber
\bm{G}=\left[
	\begin{array}{lcr}
	-1&~\bm{0}_{6}^{\top}& \\
	\bm{0}_{n}&~-\bm{\Omega}_{I}^{1/2}\bm{J}_{I}& \\
	0&~\bm{0}_{6}^{\top}& \\
	\bm{0}_{n}&~-\bm{\Omega}_{D}^{1/2}\bm{J}_{D}&
	\end{array}
	\right],~
\bm{h}=\left[
	\begin{array}{lcr}
	0 \\
	\bm{\Omega}_{I}^{1/2}\bm{r}_{I} \\
	\sqrt{\epsilon_{D}} \\
	\bm{\Omega}_{D}^{1/2}\bm{r}_{D}
	\end{array}
	\right],
\end{equation}
where $\bm{0}_{n}$ denotes the zero column vector in $\mathbb{R}^{n}$.

The upper bound $\epsilon_{D}$ of the depth objective $F_{D}(\bm{\xi})$ is a parameter that needs to be selected before starting the optimization procedure. This parameter plays the same role as $\lambda$ in \eqref{new weighted sum form}, i.e., balancing the relative importance of the depth and intensity objectives. However, compared to $\lambda$, the upper bound $\epsilon_{D}$ has a more explicit mathematical meaning and is easier to select a priori. In fact, since the value of $F_{D}(\bm{\xi})$ can be measured directly when the true value of the camera motion $\bm{\xi}^{*}$ is plugged into $F_{D}(\bm{\xi})$, it can be used to estimate the range of $\epsilon_{D}$ and find a good $\epsilon_{D}$ for optimization. In our algorithm, we choose the value of $\epsilon_{D}$ according to the complexity of depth image as follow:
\begin{equation}\nonumber
	\epsilon_{D}=
	\begin{cases}
	\epsilon_{\text{max}},&\text{if $\pi(D) \leq \delta$},\\
	\epsilon_{\text{min}},&\text{otherwise},
	\end{cases}
\end{equation}
where $\epsilon_{\text{min}} \ll \epsilon_{\text{max}}$, $\delta$ is an adjustable threshold and $\pi(D)$ is the depth metric in \eqref{depth complexity metric}.

%==========================================================================================
\section{PERFORMANCE EVALUATION}\label{evaluation}

\begin{table*}
\centering
\caption{RMSE result for the 1st-4th sequences in ``Structure vs. Texture'' category. In these sequences, the TGB-D camera is close to the panels and wooden surfaces.}
\label{table:1}
\begin{tabular}{c|c|c|c|c}
\hhline
                 & poor structure    & rich structure    & poor structure    & rich structure     \\
Method           & rich texture      & poor texture      & poor texture      & rich texture       \\
                 & [m/s]             & [m/s]             & [m/s]             & [m/s]              \\
\hline
Single objective & 0.041667          & 0.125235          & 0.249357          & 0.015956           \\ \hline
Tykkala's method & 0.035970          & 0.106649          & \textbf{0.165702} & 0.016078           \\ \hline
Weighted sum     & 0.034464          & \textbf{0.088853} & 0.178571          & \textbf{0.015101}  \\ \hline
Bounded objective& \textbf{0.032715} & 0.095749          & 0.178994          & 0.015330           \\
\hhline
\end{tabular}
\end{table*}

\begin{table*}
\centering
\caption{RMSE result for the 5th-8th sequences in ``Structure vs. Texture'' category. In these sequences, the TGB-D camera is far from the panels and wooden surfaces.}
\label{table:2}
\begin{tabular}{c|c|c|c|c}
\hhline
                 & poor structure    & rich structure    & poor structure    & rich structure     \\
Method           & rich texture      & poor texture      & poor texture      & rich texture       \\
                 & [m/s]             & [m/s]             & [m/s]             & [m/s]              \\
\hline
Single objective & 0.110646          & 0.074372          & 0.170460          & 0.015597           \\ \hline
Tykkala's method & 0.094845          & 0.077504          & 0.129923          & 0.014728           \\ \hline
Weighted sum     & \textbf{0.078033} & 0.076853          & \textbf{0.123848} & \textbf{0.014284}  \\ \hline
Bounded objective& 0.098715          & \textbf{0.066008} & 0.152104          & 0.015269           \\
\hhline
\end{tabular}
\end{table*}

For performance evaluation, we conducted a series of numerical experiments in which a set of image sequences were used to compute simulated camera trajectory. The image sequences used in our experiments are from ``Structure vs. Texture'' category in the TUM RGB-D dataset. Images in this category can be demonstrated four different types as shown in Fig. 4. The image sequences in this dataset were created using colorful plastic foils to create textural features and white plastic foils to decrease textural features. Similarly, zig-zag structure built from wooden panels are used to increase the structural features in images while planar surfaces are used to make the strucure features of images become poor.

We compare the estimated trajectories produced via the optimization procedures with the true trajectories and calculated the root mean square error (RMSE) of the drift in meters per second. Other RGB-D odometry methods, such as the single objective method in \cite{kerl_13icra} and a re-implementation of the bi-objective odometry in \cite{tykkala2011direct}, have also been applied in our experiments as references of our methods. Besides, we measure the runtime of different approaches on a ThinkPad E431 laptop with dual-core Intel i5-3210M CPU (2.50GHz) and 4 GB RAM to evaluate their real-time performance.

Specially, to ensure identical experimental conditions, we build the $t$-distribution model mentioned in \cite{kerl_13icra} to eliminate the outliers in data and constructed the weighting matrix in objective function for all methods we evaluated. The results of our experiments are given in Tab. \ref{table:1} and Tab. \ref{table:2} (the result of per-frame translational errors is also demonstrated in Fig. 5). It can be seen that the RMSEs of the single-objective optimization based method increase considerably when textural feature of the sequences is poor. Compared to the method based on single-objective optimization, our methods, the weighted sum method and the bounded objective method, give better performance, especially in poor textural feature cases. Tykkala's method, which also uses bi-optimization optimization, has a similar performance to ours in most cases. Our conclusion is that the new bi-objective optimization formulation for RGB-D odometry can alleviate the optimization problem become non-convex and improve the accuracy of the estimates.

\begin{figure*}
\centering
\subfigure[Translational errors of the methods in sequence with rich textural feature]{
\label{fig:5_sub:1}
\includegraphics[width=0.44\textwidth]{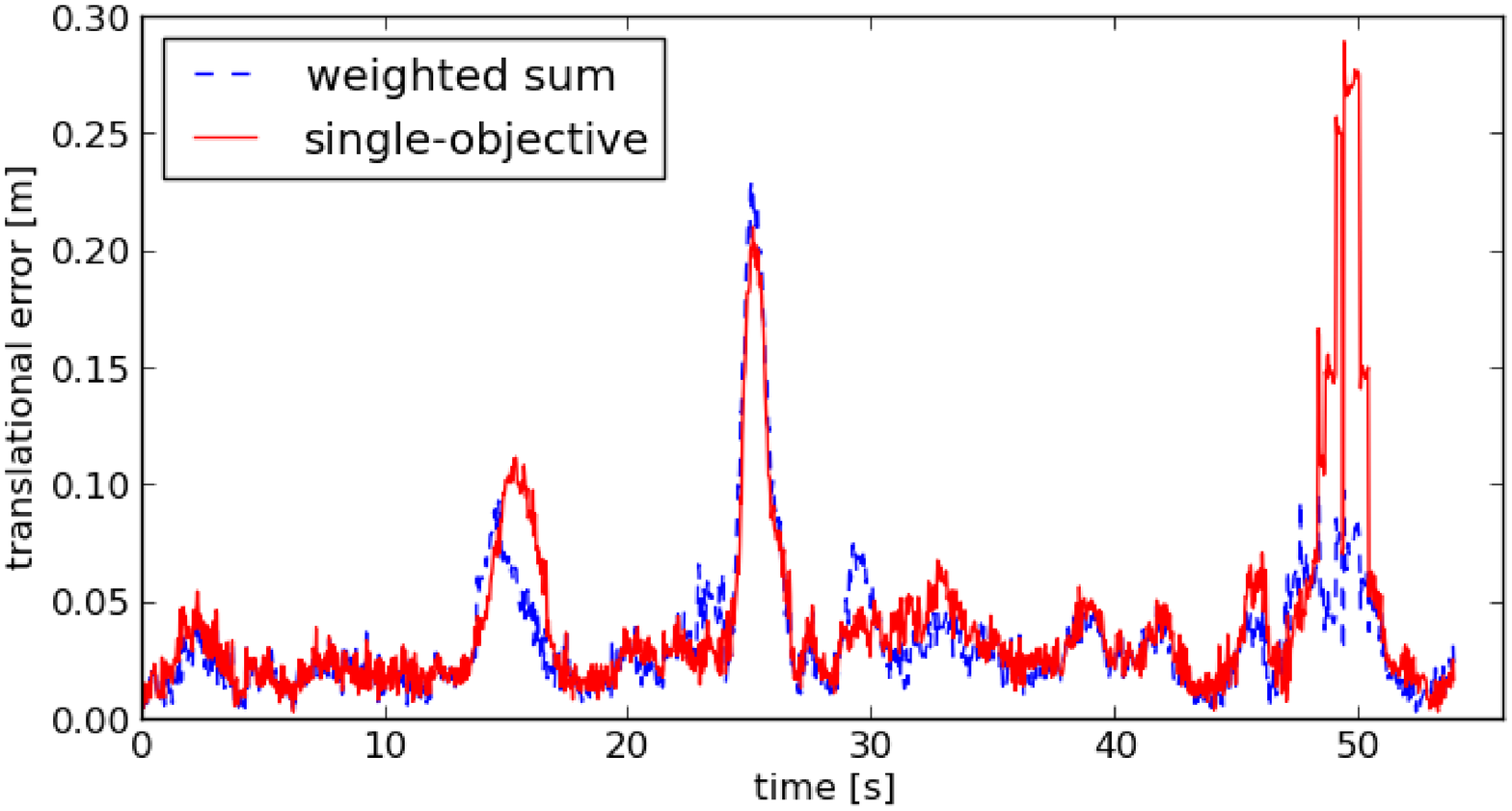}}
\subfigure[Translational errors of the methods in sequence with rich textural feature]{
\label{fig:5_sub:2}
\includegraphics[width=0.44\textwidth]{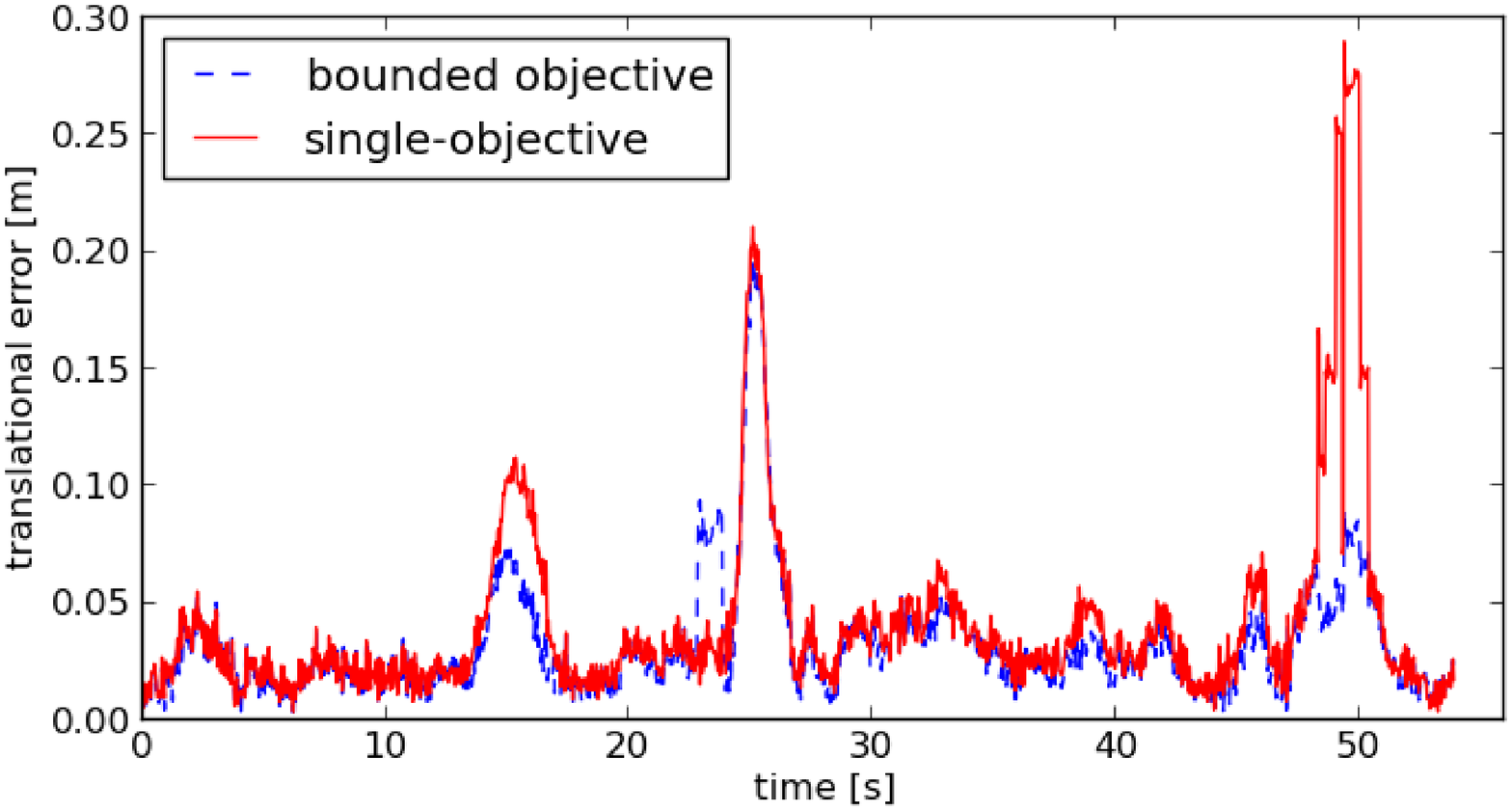}}
\subfigure[Translational errors of the methods in sequence with poor textural feature]{
\label{fig:5_sub:3}
\includegraphics[width=0.44\textwidth]{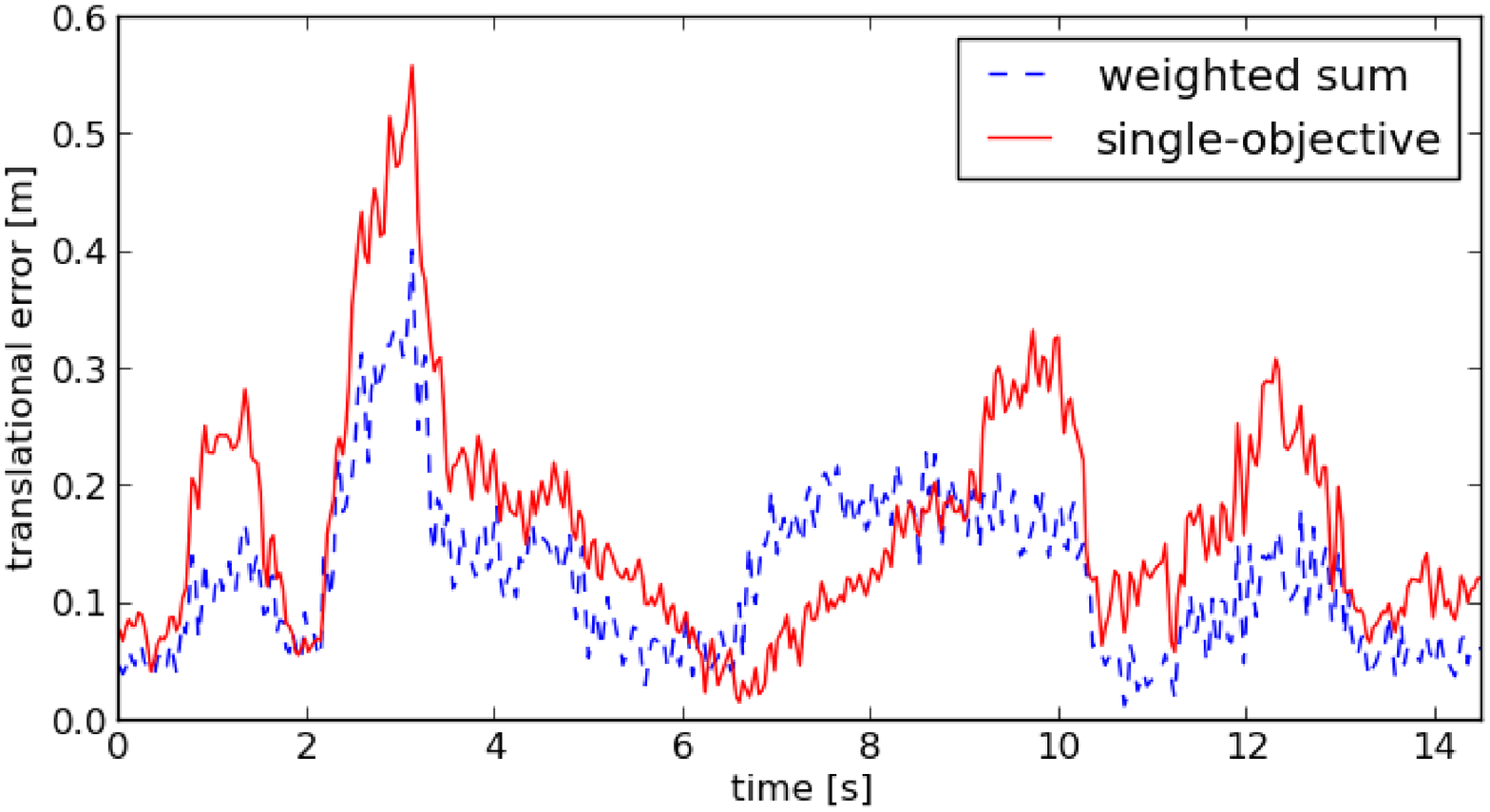}}
\subfigure[Translational errors of the methods in sequence with poor textural feature]{
\label{fig:5_sub:4}
\includegraphics[width=0.44\textwidth]{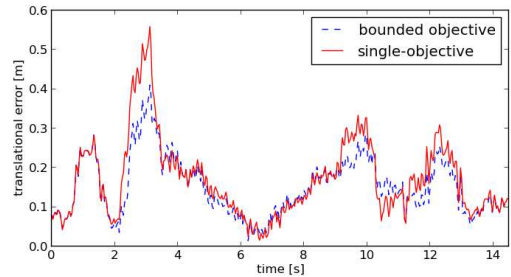}}
\caption{A comparison of per-frame translational errors between our two methods and the single objective optimization based method.}
\label{fig:5}
\end{figure*}

We also measure the average runtime for one match between two images with different methods in our experiments. From Tab. \ref{table:3} we can see that our weighted sum method needs $49.09\%$ more time to accomplish one match than the method based on single objective optimization. But as its cost in time for one match is much less than one second, our weighted sum method can still be implemented as a real-time approach. The bounded sum method, however, due to its expensive cost in time, can not work in a real-time application currently. The main cause that give rise to this phenomenon is that the algorithms used to solve the SOCP are numerical approximation algorithms. They need more computations and iterations to get the solution than the analytic algorithms, like Gauss-Newton algorithm, used in the weighted sum method. Considering its convenience in setting parameter, the bounded sum method is still a promising method and it offers an alternative beyond other common methods in bi-objecitve optimization.

\begin{table}[!htb]
\centering
\caption{Runtime result of different RGB-D odometry methods in our experiments.}\label{table:3}
\begin{tabular}{c|c}
\hhline
Method           & runtime[ms] \\ \hline
single objective & 15.42       \\ \hline
Tykkala's method & 21.06       \\ \hline
weighted sum     & 22.99       \\ \hline
bounded objective& 7093        \\
\hhline
\end{tabular}
\end{table}

%======================================================================================
\section{CONCLUSION}
In this paper, we studied two methods for solving a new bi-objective optimization formulation for robust RGB-D odometry. Both methods involve converting the bi-objective optimization problem into a single-objective problem. The weighted sum method involves minimizing the weighted linear sum of intensity and depth residuals. The bounded objective method involves minimizing the intensity residual subject to a bound on the depth residual. The experimental results show that both methods yield precise motion estimates and perform stably even when the textural information in the image sequence is poor. The bounded objective method is considerably slower than the weighted sum method. Thus, our current focus is on developing a parallel algorithm for enhancing real-time performance. We also hope to expand these ideas to other problems in robotics such as motion control, SLAM and navigation. One of the main contributions of our work is a discussion of how to use depth and intensity metrics to choose the parameters in both methods.

%======================================================================================

\balance

\end{document}